\documentclass[11pt,a4paper]{article}
\usepackage[hyperref]{emnlp2020}
\usepackage{times}
\usepackage{latexsym}

\usepackage{microtype}

\aclfinalcopy 

\usepackage{graphicx}
\usepackage{comment}
\usepackage{color, colortbl}
\usepackage{xcolor}
\usepackage{multirow}
\usepackage{adjustbox}
\usepackage{caption}
\usepackage{subcaption}

\title{BERT's output layer recognizes all hidden layers? \\ Some Intriguing Phenomena and a simple way to boost BERT}

\author{Wei-Tsung Kao\And
  Tsung-Han Wu\And Po-Han Chi\\
  National Taiwan University\\ \{\texttt{b05901009, r07942145, r08942074, r07942150, hungyilee}\}@ntu.edu.tw\And
  Chun-Cheng Hsieh\And
  Hung-Yi Lee
  }

\date{}

\begin{document}
\maketitle
\begin{abstract}
Although Bidirectional Encoder Representations from Transformers (BERT) have achieved tremendous success in many natural language processing (NLP) tasks, it remains a \textit{black box}.
A variety of previous works have tried to lift the veil of BERT and understand each layer's functionality.
In this paper, we found that surprisingly the output layer of BERT can reconstruct the input sentence by directly taking \textit{each} layer of BERT as input, even though the output layer has never seen the input other than the final hidden layer. 
This fact remains true across a wide variety of BERT-based models, even when some layers are duplicated.
Based on this observation, we propose a quite simple method to boost the performance of BERT.
By duplicating some layers in the BERT-based models to make it deeper (no extra training required in this step), they obtain better performance in the downstream tasks after fine-tuning.
\end{abstract}

\section{Introduction}

The progress of NLP based on deep learning advanced rapidly in recent years. Recently, the idea of contextualized word embeddings \citep{DBLP:journals/corr/abs-1802-05365} arises due to its better performance than static word embeddings.  
BERT~\citep{bert} pre-trains a masked language model (MLM) based on the transformer~\citep{Attention} to learn bidirectional contextualized representations of words. It can be quickly fine-tuned on many downstream tasks merely by appending a simple linear classifier, obtaining state-of-the-art results on a great number of NLP tasks. 
Although those BERT-based models can reach extraordinary performance%
, the black-box nature of deep learning models makes a fine-grained analysis on the contextualized word embedding and hidden representations much more difficult. It remains unclear that what those BERT-based models have learned and what features they extract.  

In the past, people use \textit{probing} task~\citep{context_probe} to analysis what kinds of information is contained in the hidden states, such as syntax tree~\citep{structural}, syntax and semantic in different layers~\citep{pipeline}, and word identity~\citep{Identifiability}. Although the probing task changes the explanation problem to supervised learning problem, the result heavily depends on the choice of the model family of probing model. It is not easy to separate the effect from the probing model itself \citep{designing}. 


In this paper, we found that when taking the hidden states other than the final ones as an input of the pre-trained output layer (MLM head) of BERT, it can almost reconstruct the input token sequence. The results are counterintuitive because during the pre-training stage, the output layer has never seen the input other than the final hidden states. 
This phenomenon is observed in most layers across variants of BERT-based models, including ALBERT~\citep{albert}.
This observation may lead to a whole new perspective for probing and analyzing the hidden states of BERT. 

Furthermore, the observation indicates that the hidden states of most BERT layers may have very close distribution, so they can be decoded into the input tokens by the same output layer. Therefore, we assume that duplicating layers in the model would not largely change the behavior of the model. The experimental results indeed show that when duplicating a few layers in the pre-trained BERT-based models, the input sentence can still be reconstructed by the output layer from the hidden states of almost every layer. 

\begin{figure}[th]
    \centering
    \includegraphics[width=0.95\linewidth]{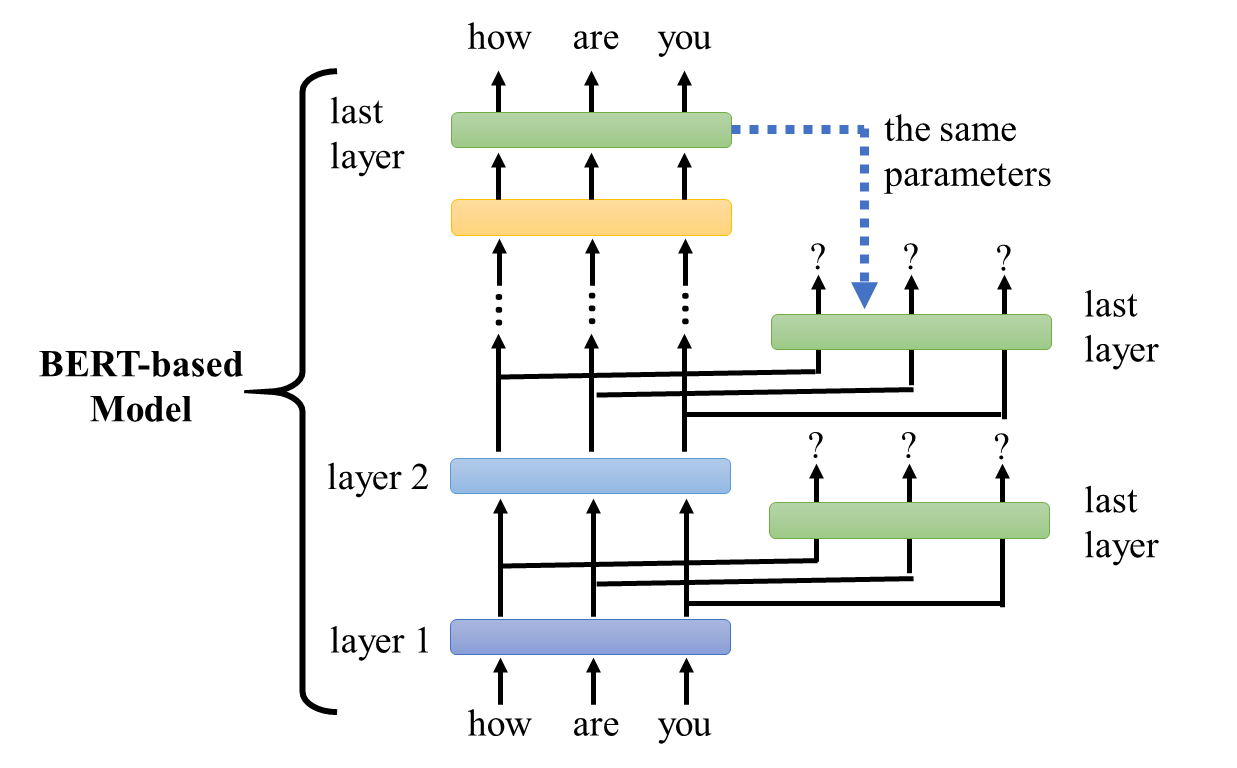}
    \caption{\textbf{Probing} hidden representations of BERT by the output layer. }
    \label{fig:decode_example}
\end{figure}

Because of the properties that duplicating layers do not change the behavior of BERT
, we can simply use this approach to make the BERT-based models deeper without any extra effort.
Then, we fine-tune the deeper models on several benchmark NLP datasets as downstream tasks 
and find that this simple duplicating layer trick can sometimes boost the model performance. 


\section{\textit{Probing} by Output Layer}
\label{sec:Decode}

\subsection{Approach}

The approach we used to \textit{probe} the BERT-based models is shown in Figure~\ref{fig:decode_example}.
During the pre-training stage, the model takes a token sequence with masking as input and learns to reconstruct the original unmasked sequence.
The output layer of the pre-trained BERT is a classifier with one hidden layer, which takes the hidden states of the last layer as input, and then output tokens. 
Here we used this output layer as a decoder to probe the hidden states of BERT.
Given an input token sequence, each layer of BERT outputs a sequence of vectors (hidden states). 
The output layer transforms each vector back into a token. 
Then we use the decoded token sequence to analyze behavior of each layer of BERT.
Note that the output layer has never seen the hidden states other than the last hidden layer during pre-training; nevertheless, when we use this output layer to decode every hidden layer, it does reconstruct the input to some good extent.
Our approach is quite different from the typical probing approach. 
In a typical approach, it requires extra training data to train the probe (a classifier).  
As for our method, we use the pre-trained output layer as a decoder and no training is required. 


\subsection{Experiment}
\label{exp:decode}

\begin{figure}[t]
    \centering
    \includegraphics[width = 0.45\textwidth]{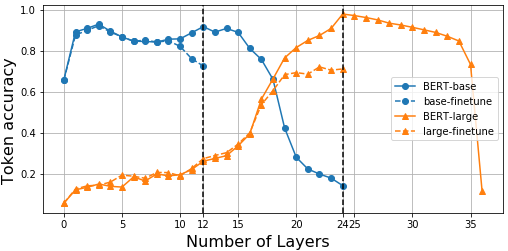}
    \caption{Average token accuracy of BERT on SST-2 dataset. layer 0 means the static word embedding layer. }
    \label{fig:sst_bert_acc}
\end{figure}

\begin{figure*}[th]
    \centering
    \includegraphics[width = 0.85\textwidth]{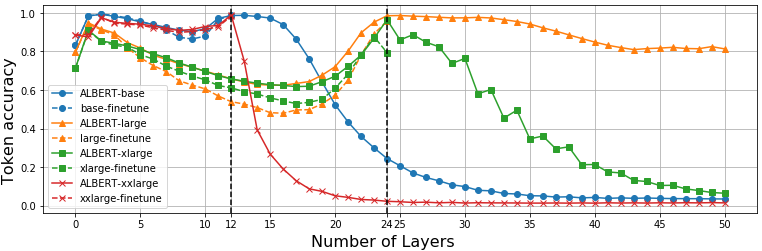}
    \caption{Average token accuracy of ALBERT on SST-2 dataset. layer 0 means the static word embedding layer. }
    \label{fig:sst_albert_acc}
\end{figure*}
\definecolor{Gray}{gray}{0.9}

\begin{table}[t]
    \centering
    \begin{adjustbox}{width=0.48\textwidth}
    \begin{tabular}{c|l}
    \hline
         Layer   & Example of Decoded Sentence \\
         \hline
         Input   & it's a bittersweet and lyrical mix of elements.\\
         \hline
         \rowcolor{Gray}
          0 & \#\#ningtonme s a bittersweettrix lyrical mixfine elements.\\
          \rowcolor{Gray}
          1 & itme s a bittersweetckle lyrical mix of elements, \\ 
         \rowcolor{Gray}
          2 & itist s a bittersweet and lyrical mix of elements, \\ 
          3 & it was s a bittersweet and lyrical mix of elements. \\
         \rowcolor{Gray}
          4 & it was s aconsweet and lyrical mix of elements. \\
          5 & it was s a bittersweet and lyrical mix of elements. \\
         \rowcolor{Gray}
          6 & it was was a bittersweet and lyrical mix of elements.\\
          \rowcolor{Gray}
          7 & it was was a bittersweet and lyrical mix of elements.\\
         \rowcolor{Gray}
          8 & it's a bittersweet and lyrical mix of elements.\\
          9 & it's a bittersweet and lyrical mix of elements.\\
         \rowcolor{Gray}
          10 & it's a bitter souleet and lyrical mix of elements.\\
          11 & album's a bitter souleet and lyrical mix of elements.\\
         \rowcolor{Gray}
          12 & .'s a sadseet and lyrical mix of elements. \\
         \hline
    \end{tabular}
    \end{adjustbox}
    \caption{Examples of probing using the output layer. }
    \label{tab:decode_example}
\end{table}

For all models in our experiments, we use the pre-trained weight from transformers library~\citep{Wolf2019HuggingFacesTS}. We use an uncased version of BERT and version-2 ALBERT model. 
Also, in the following experiments, we only analyze the input sentences without masking. The reason is that when we use the BERT models in the downstream tasks, all input sequences are not masked. This is the exact behavior we really interested in.

By considering the input token sequence as the ground truth, we compute token level accuracy for each layer. The curve at the left-hand side of the black vertical dash line of Figure~\ref{fig:sst_bert_acc} and Figure~\ref{fig:sst_albert_acc} shows average token level accuracy of BERT and ALBERT models, with input sentence from SST-2 dataset. 
The model can be either find-tuned on the downstream task (SST-2) or not.
Results on SNLI and SQuAD are similar and can be found in the Appendix A. 
Surprisingly, the accuracy of the intermediate layers of most models is higher than 80\% except layer 0 (static embeddings) and BERT-large.
That is, the input sentence can be reconstructed pretty well from all the layers by the output layer even though it only sees the last hidden layer during training. We can also see that fine-tuning only affects the last few layers, while other layers remain intact. 
Table~\ref{tab:decode_example} provides an example of probing. 

The reason for this reconstruction phenomenon is still under investigation.
One may concern that the phenomenon comes from nothing but the skip-connection or LayerNorm structure in BERT-based models. 
However, this cannot explain why BERT-large and BERT-base have such a large gap between their accuracy, especially in former layers, since both have these two structures.
It is hard to claim that successful reconstruction can be guaranteed merely with these network structures. 


The above experiments indicate that each layer does not change the input much, so most of the layers can be successfully decoded by the same output layer decoder.
Consequently, we assume that duplicating layers would not have a great influence on models.
To verify this, we examine whether the input sentence can be reconstructed by the output layer if we duplicate some layers to make it deeper than the original pre-trained ones.


The results are also shown in Figure~\ref{fig:sst_bert_acc} and Figure~\ref{fig:sst_albert_acc}. Each point after the black vertical dash line is the token accuracy of the output layer taking the last hidden states of the model with duplicated layer as input (detail of duplicating layers is in the next section.). 
The accuracy remains high for most models with duplicated layers except ALBERT-xxlarge. Even when we duplicate ALBERT-large to 50 layers, the accuracy is still about 80\%. 

Based on these results, we conclude that duplicating some layers without training may not change the pre-trained model's behavior, especially for ALBERT-large. As a result, we have the experiments in the next section.

\begin{figure}[t]
    \centering
    \includegraphics[width = 0.4\textwidth]{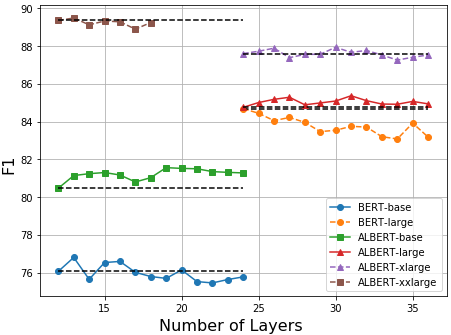}
    \caption{F1 on SQuAD 2.0 dev set. For the same curve, every point on it represents one model with duplicated layers. The leftmost point of the curve and the horizontal dotted line are the baseline model performance.}
    \label{fig:exp_f1}
\end{figure}

\begin{figure}[t]
    \centering
    \includegraphics[width = 0.4\textwidth]{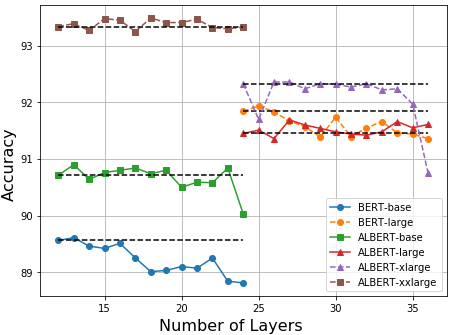}
    \caption{Accuracy on SNLI test set.}
    \label{fig:exp_snli}
\end{figure}

\section{Duplicating Layers in Downstream Tasks}
\label{sec:app_duplicate}

\subsection{Approach}

In this subsection, we propose a new method to boost the performance of BERT-base models.
The basic idea is that we duplicate some layers in the BERT before fine-tuning, make the model deeper, and fine-tune the model with more layers on the downstream tasks. 
We found that this approach can slightly improve the performance of the BERT-based models after fine-tuning with little cost.

For BERT, if we want to make the 12-layer BERT-base model become 15-layer, we duplicate the first three layers and insert each of them after the original one to achieve that; each layer will only be duplicated once in our experiment. 
In this way, the 12-layer BERT-base model can only be extended to 24-layer at most. Besides, during fine-tuning we do not tie weights of the duplicated layer with its original layer. 
If we extend a 12-layer BERT into 24-layer BERT, there should be 24 different layers of weights in the model after fine-tuning.

For ALBERT, since all layers share the same set of parameters, we do not need to consider where to insert new duplicated layers; instead, we only have to determine how many layers we want it to be. For example, to duplicate 12-layer ALBERT-base to 15 layers, we just need to apply the same parameters 15 times. Then we fine-tune these models with more layers to see whether its performance can be improved.


\subsection{Experiment}
\setlength{\tabcolsep}{2pt}
\label{exp:duplicate}

Here we show the results of duplicating layers in BERT-based models after fine-tuning. 
For the downstream tasks, we use SST-2~\citep{GLUE}, SNLI~\citep{snli:emnlp2015}, and SQuAD 2.0~\citep{squad2}.
When fine-tuning on the downstream tasks, we carefully chose hyperparameters to make sure the baseline models without duplicated layers reach the performance close to the state-of-the-art.
When fine-tuning those models with duplicated layers, we merely used the same hyperparameters as their baseline models'. 
Therefore, it is worthwhile to mention that for all models with duplicated layers in this paper, \textit{we do not choose any hyperparameters for them}. 
It is very likely that the proposed approach can show even better performance if we carefully choose the hyperparameters.
Also, we trained each baseline model three times and average to show more robust results. 

The results of single model with different numbers of duplicated layers are shown in figure~\ref{fig:exp_f1} and~\ref{fig:exp_snli}. Interestingly, many models with duplicated layers have higher performance than the baseline models. The model with duplicated layers can boost about $1$ F1 score \footnote{The same model also boosts 1.5 EM score} on SQuAD 2.0, and $0.6\%$ accuracy on SNLI. That is, directly duplicating layers on pre-trained BERT-based models before fine-tuning does not ruin the whole model; instead, it improves the model performance. 

The results of SST-2 are not shown since we found that models with duplicated layers can hardly obtain improvement on SST-2. We left the result in Appendix B.  The possible reason is that SST-2 is relatively simple. Thus, the vanilla models already reach quite high performance, leaving little room for improvement for those deeper models.

\begin{table}[t]
\centering

\begin{tabular}{c|ll}
\hline
 \multirow{2}{*}{Ensemble Models}  &\multicolumn{2}{c}{Squad 2.0}\\
    &EM&F1\\
\hline
 \rowcolor{lightgray}
 
 BERT-base   & 74.04      &  77.01 \\
- with duplicated layer (13,15,16)      & 74.91  & 77.86 \\
\hline
\rowcolor{lightgray}
ALBERT-base   & 78.85      &  82.04   \\
- with duplicated layer (19,20,21) & 79.01     &  82.02     \\
 \rowcolor{lightgray}
ALBERT-large    
& 82.49     &  85.63    \\
 - with duplicated layer (27,31,32)  & 82.73    &  85.87     \\
 \rowcolor{lightgray}
 ALBERT-xlarge    & 84.97     &  88.06  \\
- with duplicated layer (26,30,32)  & 85.50    &  88.59     \\
 \rowcolor{lightgray}
 ALBERT-xxlarge    & 86.85    &  89.99    \\
 - with duplicated layer (13,15) & 86.87 & 89.91   \\
\hline
\end{tabular}
\caption{Performance of BERT ensemble models. For baseline model, we ensemble three models trained with different random seeds. The number in the parentheses means number of layers. For ALBERT-xxlarge, only two better models are used.}
\label{tab:bert_ens}
\end{table}

Additionally, we try to ensemble top-3 models with different duplicated layers, hoping model with different layers can further boost the ensemble method. We do the ensemble experiments on SQuAD 2.0, which is the most difficult one among all tasks in this work.
Our ensemble method is merely summing up all the output probabilities of every model. In Table~\ref{tab:bert_ens} we can see that the ensemble of those models with duplicated layers boost the performance. 

\section{Conclusion and future work}
In this paper, we propose a brand new way to analyze the representations from different layers of the BERT-based models.
We observe that the representations from almost every layer can be reconstructed to the input sentences by the output layer in BERT-base and ALBERT, even though it has never seen these representations during training. 
The counterintuitive results are intriguing, which provides another viewpoint when trying to analyzing BERT-based models, and will make the researchers rethink what the BERT-based models learn in each layer.
Last but not least, we propose a pretty simple method, duplicating layers, to improve models. On SQuAD 2.0 and SNLI, our method can boost those models with state-of-the-art performance almost effortlessly. This trick can be further utilized to improve the ensemble methods as well.

\bibliographystyle{acl_natbib}
\bibliography{ref}

\newpage

\appendix

\section{Probing by output layer result on other datasets}
\label{sec:appendix}

\begin{figure}[th]
    \begin{subfigure}[t]{\columnwidth}
        \includegraphics[width=0.98\linewidth]{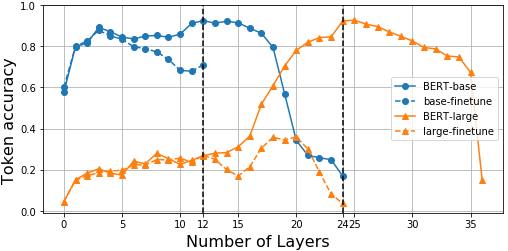}
        \caption{SNLI}
        \label{fig:snli_bert_acc}
    \end{subfigure}
    \begin{subfigure}[t]{\columnwidth}
        \includegraphics[width=0.98\linewidth]{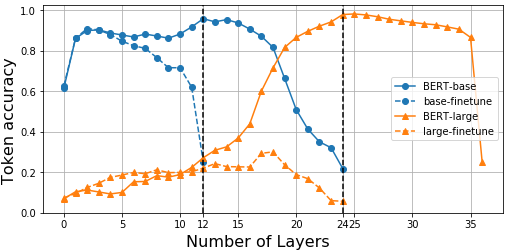}
        \caption{SQuAD 2.0}
    \label{fig:squad_bert_acc}
    \end{subfigure}
    \caption{Average token accuracy of BERT on SNLI and SQuAD 2.0. Layer 0 means the static word embedding layer. }
\end{figure}

\begin{figure*}[th]
    \centering
    \begin{subfigure}{2\columnwidth}
        \includegraphics[width=\linewidth]{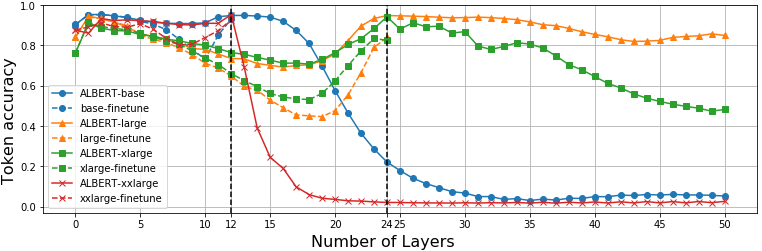}
        \caption{SNLI}
        \label{fig:snli_albert_acc}
    \end{subfigure}
    \centering
    \begin{subfigure}{2\columnwidth}
        \includegraphics[width = \linewidth]{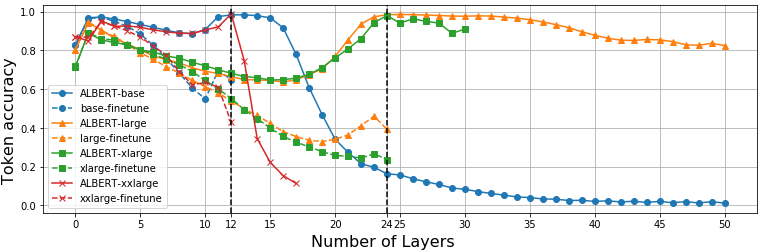}
        \caption{SQuAD 2.0}
    \label{fig:squad_albert_acc}
    \end{subfigure}
    \caption{Average token accuracy of ALBERT on SNLI and SQuAD 2.0. Layer 0 means the static word embedding layer. }
\end{figure*}

Here we show the probing result of the other two datasets, SNLI and SQuAD 2.0. For BERT, the results are in Figure~\ref{fig:snli_bert_acc} and Figure~\ref{fig:squad_bert_acc}; as for ALBERT, the results are in Figure~\ref{fig:snli_albert_acc} and Figure~\ref{fig:squad_albert_acc}. The curves at the left-hand side of the black vertical lines are the results of the pre-trained models and the fine-tuned models. On the other hand, the ones at the right-hand side of the black vertical lines are the results of those models with duplicated layers. Due to computational limitation, we can not duplicate ALBERT-xlarge and ALBERT-xxlarge to 50 layers on SQuAD 2.0, whose input sequences are much longer. 

For the pre-trained models, the reconstruction accuracy of each layer remains high over all datasets. For fine-tuned BERT and ALBERT models on SNLI and SQuAD, only the last few layers are affected, which is similar to the result on SST-2. 

For the models with duplicated layers, the results of SNLI and SQuAD 2.0 are similar to SST-2. The accuracy remains high after adding several duplicated layers, except for ALBERT-xxlarge. For ALBERT-large, the accuracy is still 80\% when duplicating to 50 layers. These results show that the phenomenon is quite general and not depends on some specific dataset.

\section{Duplicating Layers result on SST-2}

\begin{figure}[h]
    \centering
    \includegraphics[width = 0.45\textwidth]{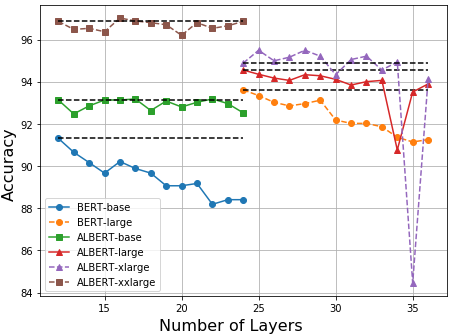}
    \caption{Accuracy of fine-tuned models on SST-2.}
    \label{fig:exp_sst}
\end{figure}

The result of duplicating layer and fine-tuning on SST-2 dataset is in Figure~\ref{fig:exp_sst}. Comparing to SNLI, the improvement on SST-2 is relatively small. Only ALBERT-xlarge and -xxlarge can be improved by duplicating layers. This result may come from that SST-2 is much simpler than SNLI and SQuAD 2.0, and can not benefit from adding more layers in the model. 

\section{Statistics of datasets}

\begin{table}[ht]
\centering

\begin{tabular}{c|lll}
\hline
  dataset  & train & dev & test \\
\hline
  SST-2 &  6920 & 872 & 1821 \\
\hline
  SNLI & 550,152 & 10,000 & 10,000 \\
\hline
  SQuAD 2.0 & 130,319 & 11,873 & 8,862 \\ 
\hline
\end{tabular}
\caption{train/dev/test examples of SST-2, SNLI, and SQuAD 2.0}
\label{tab:data_stat}
\end{table}

Table~\ref{tab:data_stat} shows the statistics of datasets used as downstream tasks. For SST-2, we use the preprocessing from \url{https://github.com/AcademiaSinicaNLPLab/sentiment_dataset}, which collects all sentence level examples in the dataset, similar to GLUE dataset. 

For SNLI, the number in Table~\ref{tab:data_stat} stands for the number of pairs of sentences. SNLI dataset can be downloaded from \url{https://nlp.stanford.edu/projects/snli/}. No data is excluded.

Lastly, for SQuAD 2.0, we truncate the sequence length of examples to 384 if it exceeds. The training set and the development set can be downloaded from \url{https://rajpurkar.github.io/SQuAD-explorer/}.

\end{document}